# UNIVERSAL AND DETERMINED CONSTRUCTORS OF MULTISETS OF OBJECTS

## Dmytro Terletskyi

***Abstract***: *This paper contains analysis of creation of sets and multisets as an approach for modeling of some aspects of human thinking. The creation of sets is considered within constructive object-oriented version of set theory (COOST), from different sides, in particular classical set theory, object-oriented programming (OOP) and development of intelligent information systems (IIS). The main feature of COOST in contrast to other versions of set theory is an opportunity to describe essences of objects more precisely, using their properties and methods, which can be applied to them. That is why this version of set theory is object-oriented and close to OOP. Within COOST, the author proposes universal constructor of multisets of objects that gives us a possibility to create arbitrary multisets of objects. In addition, a few determined constructors of multisets of objects, which allow creating multisets, using strictly defined schemas, also are proposed in the paper. Such constructors are very useful in cases of very big cardinalities of multisets, because they give us an opportunity to calculate a multiplicity of each object and cardinality of multiset before its creation. The proposed constructors of multisets of objects allow us to model in a sense corresponding processes of human thought, that in turn give us an opportunity to develop IIS, using these tools.*

***ACM Classification Keywords***: *I.2.0 General – Cognitive simulation, F.4.1 Mathematical Logic – Set theory, D.1.5 Object-oriented Programming, D.3.3 Language Constructs and Features – Abstract data types, Classes and objects, Data types and structures, E.2 Data Storage Representations – Object representation.*

## Introduction

Nowadays there are different versions of set theory, such as naive set theory of Cantor [Cantor, 1915], type theory of Russell [Wang, Mc Naughton, 1953], Zermelo-Fraenkel set theory [Fraenkel, Bar-Hillel, 1958; Wang, Mc Naughton, 1953], Von Neumann-Bernays-Gedel set theory [Wang, Mc Naughton, 1953], systems of Quine's set theory [Wang, Mc Naughton, 1953], constructible sets of Mostowski [Mostowski, 1969], alternative set theory of Vopenka [Vopenka, 1979], etc. where definition of set is introduced in different ways. Nevertheless, these definitions just describe the concept of set, and do not explain the origin of particular sets. It means they just declare a fact of existence of sets. That is why questions about the origin of specific sets are arising. Of course, we can conclude that the "new" set can be obtained by set-theoretic operations over "existing" sets, and it is really so. However, the questions about origin of these so-called "existing" sets do not disappear, because if they exist, it means that someone, using some methods (algorithms), created them earlier.

Apart from this, concept of set has important place in human thinking activity during perception, analysis, comparison, retrieval, classification and so on. Really, let us consider situation, when you have bunch of keys and need to open certain lock. If you know how exactly corresponding key looks, you can imagine and distinguish it from other keys from this bunch. In this case, it will be easy and fast. However, into another case you need to check the keys. It means, you perform certain exhaustive search, and at the same time, you create set of keys, which you have checked. Let us imagine another situation, when you need to count money, which you have in your wallet. During counting, you create at least two sets, set of banknotes and set of coins. In addition, we can consider situation when you want to play chess or checkers, and before starting, you need to make initial arrangement of figures on the chessboard. During figures placement, you create set of white and set of black figures from set of all figures. During the game, you create set of beaten figures and set of unbeaten figures from the set of all figures. These are just a few simple examples from our daily activity. Usually we pay little attention to how do we think, and what concepts do we use during this activity. However, we operate with sets of objects permanently, sometimes it happening consciously sometimes not, but it is so. These facts give as an opportunity to conclude that set is the one of basic constructions of human thinking.

Today, we have an opportunity to use sets in programming, in particular in OOP. As a proof, there are appropriate tools within some OOP-languages for working with such data structure, in particular set in STL for C++ [Musser, Derge, Saini, 2001], HashSet, SortedSet and ISet in C# [Mukherjee, 2012], HashSet in Java [Eckel, 2006], set and frozenset in Python [Summerfield, 2010]. These tools allow sets creation, executing basic set-theoretic operations, membership checking, adding and removing of elements and checking of equivalence between sets, etc.

As we can see, concept of set is very important for mathematics and has some applications in programming, in particular OOP, as practical implementation of some aspects of mathematical set theory. However, set theory and OOP are developed separately, and opportunity to work with sets within OOP is just additional functionality of OOP. It means programmers do not develop set theory, and mathematicians do not develop implementation of set theory within programming languages, very often, these two communities have different interests. Despite this, our target is development of IIS, based on human mechanisms of information analysis, in particular, on manipulation with sets of objects, using OOP. That is why we will try to combine some ideas of set theory and OOP during design and development of such systems. We will consider some constructive version of set theory described in [Terletskyi, 2014], which is close to OOP's paradigm, and show its application for simulation of some aspects of human thinking, in particular creation of sets and multisets of objects.

## Objects and Classes

We know that each set consists of elements, which form it. Everything, phenomena of our imagination or of our world can be the elements of the set [Cantor, 1915]. From other hand, one of the main postulates of OOP is that real world is created by objects [Pecinovsky, 2013]. Combining these two ideas, we will call elements of sets – objects. Let us consider such object as natural number. It is clear that every natural number must be integer and positive. These are characteristic properties of natural numbers. It is obvious, that 2 is really a natural number, but −12 and 3.62, for example, are not natural numbers.

Let us consider another object, for instance triangle. We know that triangle is geometrical figure, which has three sides for which the triangle inequality must be satisfied. According to this, geometrical figure, which has sides 3 cm, 5 cm and 7 cm is really a triangle, but figure with sides 2 cm, 4 cm and 7 cm does not triangle. We can conclude that each object has certain properties, which define it as some essence while analyzing these facts. Furthermore, objects and their properties cannot exist separately, because if we assume the opposite, we will have contradiction. On the one hand, object cannot exist separately from its properties, because without properties we cannot imagine and cannot describe it. On the other hand, object's properties cannot exist separately from object, because without object we cannot see and cannot perceive them. That is why, we cannot consider them separately, and there are few variants of the definitions order. It means that we cannot introduce definition of object without definition of its properties and vice versa. Therefore, we decided to introduce concept of object's properties firstly.

Globally we can divide properties of objects into two types – *quantitative* and *qualitative*. We will define these two types of object's properties formally, but their semantics has intuitive nature.

**Definition 1.** Quantitative property of object $A$ is a tuple $p_i(A) = (v(p_i(A)), u(p_i(A)))$, where $i = \overline{1,n}$, $v(p_i(A))$ is an quantitative value of $p_i(A)$, and $u(p_i(A))$ are units of measure of quantitative value of $p_i(A)$

**Example 1.** Suppose we have an apple, and one of its properties is weight. We can present this property as follows $p_w(A) = (v(p_w(A)), u(p_w(A)))$, and if weight of our apple is 0.2 kg, then property $p_w(A)$ will be the following $p_w(A) = (0.2, kg)$. ♠

**Definition 2.** Two quantitative properties $p_i(A)$ and $p_j(B)$, where $i = \overline{1,n}$, $j = \overline{1,m}$, are equivalent, i.e. $Eq(p_i(A), p_j(B)) = 1$, if and only if $u(p_i(A)) = u(p_j(B))$.

**Definition 3.** Qualitative property of object $A$ is a verification function $p_i(A) = vf_i(A)$, $i = \overline{1,n}$, which defines as a mapping $vf_i(A): p_i(A) \to [0,1]$.

**Example 2.** Let us consider such object as a triangle. One of its properties is triangle inequality, which must be satisfied for its sides. We can present this property as follows $p_{ti}(T) = vf_{ti}(T)$, where $vf_{ti}(T)$ is verification function of property $p_{ti}(T)$. In this case, function $vf_{ti}(T): p_{ti}(T) \to \{0,1\}$, and it is a particular case of verification function – predicate or Boolean-valued function. ♠

We can conclude that, such approach gives an opportunity to combine description of property and its verification in the one function, i.e. verification function is a verification function and a description of property at the same time. Therefore, different algorithms can be verifiers and descriptors of properties simultaneously.

**Definition 4.** Two qualitative properties $p_i(A)$ and $p_j(B)$, where $i = \overline{1,n}$, $j = \overline{1,m}$, are equivalent, i.e. $Eq(p_i(A), p_j(B)) = 1$, if and only if $(vf_i(A) = vf_j(A)) \wedge (vf_i(B) = vf_j(B))$.

**Definition 5.** Specification of object $A$ is a vector $P(A) = (p_1(A),...,p_n(A))$, where $p_i(A)$, $i = \overline{1,n}$ is quantitative or qualitative property of object $A$.

**Definition 6.** Dimension of object $A$ is number of properties of object $A$, i.e. $D(A) = |P(A)|$.

Now, we can formulate the definition of object.

**Definition 7.** Object is a pair $A/P(A)$, where $A$ is object's identifier and $P(A)$ – specification of object.

Essentially, object is a carrier of some properties, which define it as some essence.

**Definition 8.** Two objects $A$ and $B$ are similar, if and only if $P(A) \equiv P(B)$.

In general, we can divide objects on concrete and abstract, and does not matter when or how someone created each particular object. It is material implementation of its abstract image – *a prototype*. This prototype is essentially an abstract specification for creation the future real objects. Besides properties of objects, we should allocate operations (methods) which we can apply to objects, considering the features of their specifications. Really, we can apply some *operations (methods)* to objects for their changing and for operating with them. That is why, it will be useful to define concept of object's operation (method).

**Definition 9.** Operation (method) of object $A$ is a function $f(A)$, which we can apply to object $A$ considering the features of its specification.

**Example 3.** For such objects as natural numbers $n$, $m$ we can define operations "$+$" and "$\cdot$". ♠

In OOP, programmers consider specifications and methods of objects without objects, and they call it a type or a class of objects, which consists of fields and methods [Weisfeld, 2008; Pecinovsky, 2013]. Fields of class, essentially, are specification of class. Methods are functions, which we can apply to objects of this class for their changing and for operating with them. For convenience, we will also use word "signature" for methods of class. Let us define concept of object's signature.

**Definition 10.** Signature of object $A$ is a vector $F(A) = (f_1(A),...,f_m(A))$, where $f_i(A)$, $i = \overline{1,m}$ is an operation (method) of object $A$.

Generally, signature of particular object can consist of different quantity of operations, but in practice, especially in programming, usually we are considering finite signatures of objects.

According to definition of object, every object has some specification, which defines it as some essence. There are some objects, which have similar specifications. It means that we can apply the same methods to them. Let us define similar objects.

**Definition 11.** Objects $A$ and $B$ are similar objects, if and only if, they have the same dimension and equivalent specifications.

If certain two objects are similar, we can conclude that these objects have the same type or class. Now we can introduce concept of object's class.

**Definition 12.** Object's class $T$ is a tuple $T = (P(T), F(T))$, where $P(T)$ is abstract specification of some quantity of objects, and $F(T)$ is their signature.

When we talk about class of objects, we mean properties of these objects and methods, which we can apply to them. Class of objects is a generalized form of consideration of objects and operations on them, without these objects.

**Example 4.** Let us describe type *Int* in programming language C++, using concept of similar objects and object's class. Let us set the next specification for the class $P(Int) = (p_1(Int), p_2(Int))$, where property $p_1(Int)$ means "integer number", property $p_2(Int)$ means "number not bigger then 2147336147 and not smaller than -2147336148". It is obvious, that all numbers which have properties $p_1(Int)$ and $p_2(Int)$ are objects of class Int. Let define the methods of class *Int* in the following way $F(Int) = (f_1(Int), f_2(Int))$, where $f_1(Int) = "+"$ and $f_2(Int) = "*"$. ♠

As we know, in OOP, every particular object has the same fields and behavior as its class, i.e. it has the same specification and signature. It means that every class of OOP is homogeneous in a sense. That is why, let us define concept of homogeneous class of objects.

**Definition 13.** Homogeneous class of objects $T$ is a class of objects, which contains only similar objects.

The simplest examples of homogeneous classes of objects are class of natural numbers, class of letters of English alphabet, class of colors of the rainbow, etc.

Clearly, that every object is a member of at least one class of objects. Furthermore, some objects are members of few classes simultaneously. For example, such objects as natural numbers $n_1, \ldots, n_m$ are members of such classes as natural numbers $N$, integer numbers $Z$, rational numbers $Q$ and real numbers $R$. It is obvious that, class $R$ has the biggest cardinality. Furthermore, it consists of groups of objects of different types. It contradicts concept of OO-class, because different objects from one OO-class cannot have different specifications and signatures. According to this, we cannot describe such classes of objects using concept of homogeneous class. That is why we will define concept of inhomogeneous class of objects.

**Definition 14.** Inhomogeneous class of objects $T$ is a tuple

$$T = (Core(T), pr_1(A_1), \ldots, pr_n(A_n)),$$

where $Core(T) = (P(T), F(T))$ is the core of class $T$, which includes properties and methods similar to specifications $P(A_1), \ldots, P(A_n)$ and signatures $F(A_1), \ldots, F(A_n)$ respectively and $pr_i(A_i) = (P(A_i), F(A_i))$, $i = \overline{1, n}$ is projection of object $A_i$, which consists of properties and methods typical only for this object.

The simplest examples of inhomogeneous classes of objects are class of polygons, cars, birds, etc.

**Definition 15.** Two classes of objects $T_1$ and $T_2$ are equivalent, i.e. $Eq(T_1, T_2) = 1$, if and only if $(P(T_1) \equiv P(T_2)) \wedge (F(T_1) \equiv F(T_2))$.

**Sets and Multisets of Objects**

According to Naive set theory, a set is a gathering together into a whole of definite, distinct objects of our perception or of our thought, which are called elements of the set [Cantor, 1915]. As we can see, this definition just describes concept of set, and does not explain how to gather these objects together. That is why we are going to define union operation on objects, as a method of set creation.

**Definition 16.** Union $\cup$ of $n \geq 2$ arbitrary objects is a new set of objects $S$, which is obtained in the following way

$$S = A_1 / T(A_1) \cup ... \cup A_n / T(A_n) = \{A_1,...,A_n\} / T(S),$$

where $\forall A_i, A_j \in S$, $i,j = \overline{1,n}$ and $i \neq j$, $Eq(A_i, A_j) = 0$; $T(A_i)$, $i = \overline{1,n}$ is a class of object $A_i$ and $T(S)$ is a class of new set of objects $S$ and $n$ is its cardinality.

**Example 5.** Let us consider such geometrical objects as triangle, square and trapeze. It is obvious that these objects belong to different classes of polygons. Let us denote triangle as $A$, square as $B$, trapeze as $C$, and describe their classes as follows

$$T(A) = ((p_1(A),...,p_5(A)),(f_1(A),f_2(A))); \; T(B) = ((p_1(B),...,p_4(B)),(f_1(B),f_2(B)));$$

$$T(C) = ((p_1(C),...,p_5(C)),(f_1(C),f_2(C))).$$

Properties $p_1(A)$, $p_1(B)$, $p_1(C)$ are quantities of sides of figures, properties $p_2(A)$, $p_2(B)$, $p_2(C)$, are sizes of sides of figures, properties $p_3(A)$, $p_3(B)$, $p_3(C)$ are quantities of angles of figures, properties $p_4(A)$, $p_4(B)$, $p_4(C)$ are sizes of angles of figures, property $p_5(A)$ is triangle inequality and property $p_5(C)$ is parallelism of two sides of figure. Methods $f_1(A)$, $f_1(B)$, $f_1(C)$ are functions of perimeter calculation of figures, and methods $f_2(A)$, $f_2(B)$, $f_2(C)$ are functions of area calculation of figures.

Of course, specifications and signatures of these objects can include more properties and methods, than we have presented in this example, but everything depends on level of detail. Let us define specifications and signatures of these objects (see Table 1).

Table 1. Specifications and signatures of triangle $A$, square $B$ and trapeze $C$

|   | $p_1$ | $p_2$ | $p_3$ | $p_4$ | $p_5$ | $f_1$ | $f_2$ |
|---|---|---|---|---|---|---|---|
| $A$ | 3 | 3.6 cm, 3.6 cm, 5.9 cm | 3 | 35°, 35°, 110° | 1 | $P = \sum_{i=1}^{3} a_i$ | $S = \sqrt{p(p-a)(p-b)(p-c)}$ |
| $B$ | 4 | 2 cm, 2 cm, 2 cm, 2 cm | 4 | 90°, 90°, 90°, 90° | × | $P = \sum_{i=1}^{4} a_i$ | $S = a^2$ |
| $C$ | 4 | 3.6 cm, 5.9 cm, 3.6 cm, 11.8 cm | 4 | 35°, 145°, 145°, 35° | 1 | $P = \sum_{i=1}^{4} a_i$ | $S = \frac{(a+b)}{2} h$ |

Analyzing Table 1, we can see that property $p_5$ specified as just value of verification function for particular object. All these functions can be simply implemented using OOP language. Furthermore, there are variety of their implementations that is why we will not consider them within this example.

Now, let us apply the union operation to these objects and create a new set of objects.

$$S = A / T(A) \cup B / T(B) \cup C / T(C) = \{A, B, C\} / T(S)$$

We have obtained a new set of objects $S$ and a new class of objects

$$T(S) = (Core(S), pr_1(A), pr_2(B), pr_3(C)),$$

Where $Core(S) = (p_1(S), p_2(S), p_3(S), p_4(S), f_1(S))$, property $p_1(S)$ is quantity of sides of figures, property $p_2(S)$ means sizes of sides of figures, property $p_3(S)$ is quantity of angles of figures, property $p_4(S)$ means sizes of angles of figures, method $f_1(S)$ is a function of perimeter calculation of figures,

$$pr_1(A) = (p_5(A), f_2(A)), \; pr_2(B) = (f_2(B)), \; pr_3(C) = (p_5(C), f_2(C)).$$

Essentially, the set of objects $S$ is the set of triangles of class $T(A)$, squares of class $T(B)$ and trapezes of class $T(C)$ and class of set of objects $T(S)$ describes these three types of geometrical figures. ♠

Therefore, we can create sets of object, applying union operation to objects and not only. According to classical set theory, we can do it, applying union operation to sets of objects. However, this operation does not consider concept of class of objects that is why we need to redefine it.

**Definition 17.** Union $\cup$ of $m \geq 2$ arbitrary sets of objects is a new set of objects $S$, which is obtained in the following way

$$S = S_1 / T(S_1) \cup ... \cup S_m / T(S_m) = \{A_1, ..., A_n\} / T(S),$$

where $\forall A_i, A_j \in S$, $i, j = \overline{1, n}$ and $i \neq j$, $Eq(A_i, A_j) = 0$; $T(S_i)$, $i = \overline{1, m}$ is a class of set of objects $S_i$ and $T(S)$ is a class of a new set of objects $S$ and $n$ is its cardinality.

**Example 6.** Let us consider such objects as triangle $A$, square $B$ and trapeze $C$, which belong to classes $T(A)$, $T(B)$ and $T(C)$, described in the Example 5, respectively. Let us create two sets of objects $S_1$ and $S_2$ using Definition 16, i.e.

$$S_1 = A / T(A) \cup B / T(B) = \{A, B\} / T(S_1); \; S_2 = A / T(A) \cup C / T(C) = \{A, C\} / T(S_2).$$

As the result we have obtain new sets of objects $S_1$, $S_2$ and new classes of objects $T(S_1)$, $T(S_2)$, that have following structures

$$T(S_1) = (Core(S_1), pr_1(A), pr_2(B)); \; T(S_2) = (Core(S_2), pr_1(A), pr_2(C)).$$

In the case cores of both classes are the same, it means

$$Core(S_1) = Core(S_2) = (p_1(S), p_2(S), p_3(S), p_4(S), f_1(S)),$$

where property $p_1(S)$ is quantity of sides of figures, property $p_2(S)$ means sizes of sides of figures, property $p_3(S)$ is quantity of angles of figures, property $p_4(S)$ means sizes of angles of figures, method $f_1(S)$ is a function of perimeter calculation of figures. Concerning projections of these classes, then they have following structures $pr_1(A) = (p_1(A), f_2(A))$, $pr_2(B) = (f_2(B))$, $pr_2(C) = (p_5(C), f_2(C))$.

Now, let us calculate union of $S_1$ and $S_2$.

$$S = S_1 / T(S_1) \cup S_2 / T(S_2) = \{A, B\} / T(S_1) \cup \{A, C\} / T(S_2) = \{A, B, C\} / T(S)$$

As we can see, we have obtained the same result, as in the case of union of objects $A$, $B$ and $C$, which we considered in the previous example. ♠

Consequently, we have considered two ways of set creation, however we can also obtain a set of objects, combining these two approaches.

**Definition 18.** Union $\cup$ of $n \geq 1$ arbitrary objects and $m \geq 1$ arbitrary sets of objects is a new set of objects $S$, which is obtained in the following way

$$S = A_1 / T(A_1) \cup ... \cup A_n / T(A_n) \cup S_1 / T(S_1) \cup ... \cup S_m / T(S_m) = \{A_1, ..., A_k\} / T(S),$$

where $\forall A_i, A_j \in S$, $i, j = \overline{1, k}$ and $i \neq j$, $Eq(A_i, A_j) = 0$; $T(A_v)$, $v = \overline{1, n}$ is a class of object $A_v$, $T(S_w)$, $w = \overline{1, m}$ is a class of set of objects $S_w$ and $T(S)$ is a class of new set of objects $S$ and $k$ is its cardinality.

**Example 7.** Let us consider objects $A$, $B$, $C$ and sets of objects $S_1$, $S_2$ which were described above, and calculate their union.

$$S = A / T(A) \cup B / T(B) \cup C / T(C) \cup S_1 / T(S_1) \cup S_2 / T(S_2) =$$
$$= A / T(A) \cup B / T(B) \cup C / T(C) \cup \{A, B\} / T(S_1) \cup \{A, C\} / T(S_2) = \{A, B, C\} / T(S).$$

As we can see, we have obtained the same result, as in the previous example. ♠

Let us define a concept of set of objects based on methods of set creation, which were considered above.

**Definition 19.** The set of objects $S$ is a union, which satisfies one of the following schemes:

$$S1: O_1 / T(O_1) \cup ... \cup O_n / T(O_n) = S / T(S);$$

$$S2: S_1 / T(S_1) \cup ... \cup S_m / T(S_m) = S / T(S);$$

$$S3: O_1 / T(O_1) \cup ... \cup O_n / T(O_n) \cup S_1 / T(S_1) \cup ... \cup S_m / T(S_m) = S / T(S);$$

where $O_1, ..., O_n$ are arbitrary objects, $S_1, ..., S_m$ are arbitrary sets of objects, and $T(S)$ is a class of a new set of objects $S$.

According to types of objects, which form a set of objects, we can obtain different types of sets of objects, in particular set of objects, which consists of only objects, that belong to the same class of objects.

Let us define concept of homogeneous set of objects, based on concept of homogeneous class of objects.

**Definition 20.** Set of objects $S = \{A_1, ..., A_n\}$ is homogeneous, if and only if $\forall A_i, A_j \in S$, $i, j = \overline{1, n}$ and $i \neq j$, $Eq(T(A_i), T(A_j)) = 1$.

As we know, multiset is a generalization of the notion of set in which members are allowed to appear more than once [Syropoulos, 2001]. Formally multiset can be defined as a 2-tuple $(A, m)$, where $A$ is the set, and $m$ is the function that puts a natural number, which is called the multiplicity of the element, in accordance to each element of the set $A$ i.e. $m : A \to N$. However, this definition does not explain how to create a multiset of objects that is why we are going to define multiset of objects using concept of set of objects.

**Definition 21.** The multiset of objects is a set of objects $S = \{A_1, ..., A_n\}$, such that $\exists A_i, A_j \in S$, where $i, j = \overline{1, n}$ and $i \neq j$, $Eq(A_i, A_j) = 1$.

We can obtain a multiset of objects in the same way as a set of objects.

**Example 8.** Let us consider objects $A$, $B$, $C$ and sets of objects $S_1$, $S_2$ from Example 5 and Example 6.

Union of objects.
$$S = A / T(A) \cup A / T(A) \cup B / T(B) \cup B / T(B) \cup C / T(C) = \{A, A, B, B, C\} / T(S)$$

Union of sets of objects.
$$S = S_1 / T(S_1) \cup S_2 / T(S_2) = \{A, B\} / T(S_1) \cup \{A, C\} / T(S_2) = \{A, A, B, C\} / T(S)$$

Union of objects and sets of objects.
$$S = A / T(A) \cup S_2 / T(S_2) = A / T(A) \cup \{A, C\} / T(S_2) = \{A, A, C\} / T(S)$$

Using three different ways of creation, we have obtained three different multisets of objects. ♠

Let us define some auxiliary definitions connected with multisets of objects.

**Definition 22.** Cardinality of multiset of objects $S = \{A_1, ..., A_n\}$ is a quantity of objects, which it contains, i.e. $|S| = n$.

**Definition 23.** Basic set of multiset of objects $S = \{A_1, ..., A_n\}$ is a set of objects $S_b$, which is defining as follows
$$S_b = bs(S) = \{A_1, ..., A_m\},$$
where $m \leq n$, $\forall A_i, A_j \in S_b$, $i, j = \overline{1, m}$, $i \neq j$, $Eq(A_i, A_j) = 0$ and $\forall A_w \in S$, $A_w \in S_b$.

**Example 9.** If we have set of objects $S = \{A, A, B, B, B, C, D, D\}$, then $bs(S) = S_b = \{A, B, C, D\}$. ♠

## Universal Constructor of Multisets of Objects

As we can see, a multiset of objects can be obtained similarly to sets of objects. However, sometimes we need to recognize or identify particular copy of some elements, which have multiplicity $m \geq 2$. That is why, we will consider universal constructor of multisets of objects, which was presented in [Terletskyi, 2014]. After that, we will show its generality, i.e. we can create arbitrary multiset of object, using this constructor. However, firstly we are going to define cloning operation on objects.

**Definition 24.** Clone of the arbitrary object $A$ is the object $Clone_k(A) = A_{i+k} / P(A_i)$, where $P(A)$ is a specification of object $A$, $i$ is a number of its copy and $k$ is a clone's number of $A_i$. If the object $A$ is not a clone, then $i = 0$.

The main idea of universal constructor of multisets of objects is superposition of union and cloning operation of objects.

**Example 10.** Let us consider triangle $A$ from previous section. Using cloning operation, we can create the clones of $A$, for example

$$Clone_1(A) = A_1 / (p_1(A),...,p_5(A)); \quad Clone_2(A) = A_2 / (p_1(A),...,p_5(A));$$

Clearly, that triangle $A$ and its clones $A_1$, $A_2$ are similar triangles. After this, we can apply union operation to them, and in such a way to create the multiset of triangles $S$, i.e.

$$S = A / T(A) \cup A_1 / T(A) \cup A_2 / T(A) = \{A, A_1, A_2\} / T(S).$$

Thus, when we have done it, we have also created a new class of multiset of objects $T(S)$, but in this case, it is equivalent to class $T(A)$, i.e. $P(S) = P(A) = (p_1(A),...,p_5(A))$ and $F(S) = F(A) = (f_1(A), f_2(A))$. That is why, $S$ is a homogeneous multiset of objects. ♠

Considering this example, we can conclude that

$$S = A \cup \left( \bigcup_{i=1}^{2} Clone_i(A) \right).$$

It means that we can create any multisets of objects, using arbitrary superposition of union and cloning operations of objects. According to this, we can define our universal constructor of multisets of objects (UCM) as follows

$$UCM(A, m) = A \cup \left( \bigcup_{i=1}^{m} Clone_i(A) \right),$$

where $m$ is a multiplicity of object $A$.

**Example 11.** Let us extend this constructor to inhomogeneous objects and consider for it the square $B$ and the trapeze $C$, which were defined in the previous section. Using cloning operation, we can create the clones of object $B$ and of object $C$, for example

$$Clone_1(B) = B_1 / (p_1(B),...,p_4(B)); \quad Clone_1(C) = C_1 / (p_1(C),...,p_5(C));$$

$$Clone_2(C) = C_2 / (p_1(C),...,p_5(C)).$$

Clearly, that object $B$ and its clone $B_1$ are similar squares. We have the same situation in case of trapeze $C$ and its clones $C_1$, $C_2$. After this, we can apply union operation to objects $B$, $C$ and their clones $B_1$, $C_1$, $C_2$, and in such a way to create a multiset of squares and trapezes $S$, i.e.

$$S = B / T(B) \cup B_1 / T(B) \cup C / T(C) \cup C_1 / T(C) \cup C_2 / T(C) = \{B, B_1, C, C_1, C_2\} / T(S).$$

Thus, when we have done it, we have also created a new class $T(S)$ with the following specification

$$T(S) = (Core(S), pr_1(B), pr_2(C)) = ((p_1(S), p_2(S), p_3(S), p_5(S), f_1(S)), (f_2(B)), (p_5(C), f_2(C))).$$

Clearly, that $S$ is the inhomogeneous multiset of objects, because $B \equiv B_1$ and $C \equiv C_1 \equiv C_2$, but $B$ and $C$ are objects of different classes. ♠

Considering this example, we can conclude that

$$UCM((A_1,m),...,(A_n,m_n)) = \bigcup_{i=1}^{n}\left( A_i \cup \left( \bigcup_{j=1}^{m_i} Clone_j(A_i) \right) \right),$$

where $m_1,...,m_n$ are multiplicities of objects $A_1,...,A_n$ respectively.

**Theorem 1.** *Any multiset of objects can be created using UCM.*

*Proof.* Our proof consists of two parts, in which we are going to prove that any multiset of objects can be created using UCM(1) and arbitrary multiset of objects can be reduced to input data of UCM (2).

First condition follows from UCM's definition. Really, superposition of union and cloning operation in UCM guarantees multisets of objects in the result. Type of resultant multiset of objects depends on types of objects, which are parameters for UCM and their multiplicities.

Second condition follows from that fact, that every multiset of object can be presented in accordance with the formal definition of multiset, i.e. if $S = \{A_1,...,A_k\}$ is multiset of objects, then it can be presented as follows $S = ((A_1,m_1),...,(A_n,m_n))$, where $m_1 + ... + m_n = k$, what is an input data for UCM. It means that we can create exactly the same multiset of objects, using tuple form of presentation of multiset as an input data for UCM, i.e. $UCM((A_1,m_1),...,(A_n,m_n)) = \{A_1,...,A_k\} = S$. □

As we can see, this constructor is quite general and gives us an opportunity to create different types of multisets of objects, in particular homogeneous and inhomogeneous. Clearly that this constructor is determined if and only if $m_1,...,m_n$ are strictly defined. In addition, we are going to define a few other determined constructors of multisets of objects, which strictly define the multiplicity of each element, using for it their own schema.

## CP Constructor

This constructor of multisets of objects based on the idea of Cartesian product of two arbitrary sets, that is why we call it CP constructor. We use the idea of Cartesian product of sets. However, in contrast to classical definition of CP we define pairs of CP as sets of objects.

**Example 12.** Let us consider situation that we need to construct electric garland, and we have green, yellow, orange, blue, purple and rosy light bulbs for it. Before we will make our garland, we need to create color scheme for it. It means we need to decide which colors and how many light bulbs of every color we want to use. It is convenient for us to denote every type of light bulbs according to first letter of its color. Let us assume that we want to use all colors, which we have, and each of them can be used more than once. Let us randomly divide all colors on two sets, for instance $S_1 = \{G,Y,O\}$, $S_2 = \{B,P,R\}$, and build all possible sets of objects which consist of elements of Cartesian product pairs, i.e.

$$S_1 = \{G,B\},\ S_2 = \{G,P\},\ S_3 = \{G,R\},\ S_4 = \{Y,B\},\ S_5 = \{Y,P\},$$

$$S_6 = \{Y,R\}, \ S_7 = \{O,B\}, \ S_8 = \{O,P\}, \ S_9 = \{O,R\}.$$

Let us apply union operation to these sets of objects, i.e.

$$S = \{G,B\}/T(S_1) \cup \{G,P\}/T(S_2) \cup \{G,R\}/T(S_3) \cup \{Y,B\}/T(S_4) \cup \{Y,P\}/T(S_5) \cup$$
$$\cup \{Y,R\}/T(S_6) \cup \{O,B\}/T(S_7) \cup \{O,P\}/T(S_8) \cup \{O,R\}/T(S_9) =$$
$$= \{G,B,G,P,G,R,Y,B,Y,P,Y,R,O,B,O,P,O,R\}/T(S),$$

where $S$ is multiset of objects and $T(S)$ is its class. Clearly, that all objects are similar, that is why class $T(S)$ is homogeneous. As the result, we have obtained multiset of objects, which consists of six objects $G$, $Y$, $O$, $B$, $P$, $R$ and we can consider $S$ as one of possible projects of future electric garland. ♠

Generally we can represent this scheme as follows $S = \{(G,3),(Y,3),(O,3),(B,3),(P,3),(R,3)\}$, because $S$ is a multiset of colors. Such form of presentation gives us quantity of each type of light bulbs. However, order of colors is very important aspects of garland's creation. It is obvious that different orders of the same quantity of colors and placement of particular light bulbs give us different perception of garland. According to it, we can vary different combinations of light bulbs for finding needed combination.

Sometimes we need to identify each light bulb of each color, for example for substitute. That is why we are going to improve our constructor in this aspect, via indexation operation.

**Definition 25.** Indexation of object $A_i$ is a redefining of its index $i$, i.e. $Ind(A_i) = A_{i+w} / (p_1(A),...,p_n(A))$, where $i$ is an index of object $A$ and $w$ is its increase.

According to this, the result of the Example 12 is the following

$$S = \{G_1,B_1\}/T(S_1) \cup \{G_2,P_1\}/T(S_2) \cup \{G_3,R_1\}/T(S_3) \cup \{Y_1,B_2\}/T(S_4) \cup \{Y_2,P_2\}/T(S_5) \cup$$
$$\cup \{Y_3,R_2\}/T(S_6) \cup \{O_1,B_3\}/T(S_7) \cup \{O_2,P_3\}/T(S_8) \cup \{O_3,R_3\}/T(S_9) =$$
$$= \{G_1,B_1,G_2,P_1,G_3,R_1,Y_1,B_2,Y_2,P_2,Y_3,R_2,O_1,B_3,O_2,P_3,O_3,R_3\}/T(S),$$

where $S$ is a multiset of objects and $T(S)$ is its class. According to this, we can represent our CP constructor as follows

$$CP(S_1, S_2) = \bigcup_{i=1}^{n} \bigcup_{j=1}^{m} (Ind_j(A_i) \cup Ind_i(B_j)),$$

where $S_1$, $S_2$ are basic sets of objects for multiset of objects $S$, $A_i \in S_1$, $B_j \in S_2$, $n = |S_1|$ and $m = |S_2|$.

As we can see, CP constructor gives us determined scheme for creation of multiset of objects. We also can calculate multiplicity of each object and cardinality of multiset before its creation. As a proof of these facts, we can formulate and prove following two theorems.

**Theorem 2.** *Cardinality of each multiset of objects $S$, which is obtained using CP constructor, can be calculated by the following formula*

$$|S| = 2nm,$$

where $n = |S_1|$, $m = |S_2|$.

*Proof.* As we know, cardinality of Cartesian product of two sets can be calculated as follows

$$|S_1 \times S_2| = |S_1| \cdot |S_2| = nm.$$

According to the fact, that elements of Cartesian product are pairs, we can conclude that $|CP(S_1, S_2)| = 2nm$. □

**Theorem 3.** *Multiplicity of each object $A_i$ from multiset of objects $S$, which is obtained using CP constructor, can be calculated by the following formula*

$$m(A_i) = \begin{cases} |S_2|, & \exists B_j \in S_1 \mid A_i = B_j; \\ |S_1|, & \exists C_k \in S_2 \mid A_i = C_k; \end{cases}$$

where $i = \overline{1, 2nm}$, $j = \overline{1, n}$, $k = \overline{1, m}$ and $S_1$, $S_2$ are basic sets of objects for multisets of objects $S$.

*Proof.* Proof follows from the definition of Cartesian product of sets. □

## RCL Constructor

The basic principle of this constructor is recursive cloning of set of objects that is why we call this constructor RCL constructor. We will combine the idea of object's cloning with the idea of direct recursion, within RCL constructor, but firstly we need to define cloning operation for set of objects.

**Definition 26.** Clone of the arbitrary set of objects $S = \{A_1, ..., A_n\} / T(S)$ is the set of objects

$$Clone_i(S) = \{A_{1+i}, ..., A_{n+i}\} / T(S),$$

where $T(S)$ is a class of set of objects $S$ and $i$ is a number of its clone.

**Example 13.** Let us consider Example 12 and imagine that we have only green, yellow and red light bulbs. It means, that we have set of colors $S_1 = \{G, Y, R\}$. Let us clone it once, and apply union operation to it and to the result of its cloning, i.e.

$$S_2 = S_1 / T(S_1) \cup Clone_1(S_1 / T(S_1)) = \{G, Y, R\} / T(S_1) \cup \{G_1, Y_1, R_1\} / T(S_1) =$$
$$= \{G, Y, R, G_1, Y_1, R_1\} / T(S_1).$$

As the result, we have obtained the multiset of colors $S_2$. Let us repeat the same procedure for it.

$$S_3 = S_2 / T(S_1) \cup Clone_1(S_2 / T(S_1)) = \{G, Y, R, G_1, Y_1, R_1\} / T(S_1) \cup \{G_2, Y_2, R_2, G_3, Y_3, R_3\} / T(S_1) =$$
$$= \{G, Y, R, G_1, Y_1, R_1, G_2, Y_2, R_2, G_3, Y_3, R_3\} / T(S_1).$$

where $S_3$ is a multiset of objects, and $T(S_1)$ is its class. As the result we have obtained multiset of objects which consists of three objects $G$, $Y$, $R$ and their copies, that can be accurately identified and we can consider $S$ as one of possible projects of future electric garland. ♠

Using a scheme of creation of multiset of objects from Example 12, we can represent our RCL constructor as follows

$$RCL^n(S) = \begin{cases} S, & n = 0; \\ S \cup Clone_{2^{n-1}}(S), & n = 1; \\ RCL^{n-1}(S) \cup Clone_{2^{n-1}}(RCL^{n-1}(S)), & n \geq 2. \end{cases}$$

where $S$ is a basic set of objects for multiset of objects $RCL^n(S)$ and $n$ is a recursion depth.

As we can see, RCL constructor gives us defined order of colors. We also can calculate cardinality of garland and quantity of light bulbs of each color before garland's creation. As a proof of these facts, we can formulate and prove following two theorems.

**Theorem 4.** *Cardinality of each multiset of objects $S$, which is obtained using RCL constructor, can be calculated by the following formula*

$$|S| = n2^i,$$

*where $n$ is a cardinality of basic set of objects and $i$ is recursion depth.*

*Proof.* According to the scheme of RCL constructor, on each step we will make a union of two sets of objects, which have equal cardinality. It means, if $|S_b| = n$, then on the step $i = 1$ we have a multiset of objects which cardinality is calculated as follows $n + n = 2n = n2^1$. On the step $i = 2$ we have a multiset of object with cardinality $2n + 2n = 4n$, i.e. $n2^1 + n2^1 = n2^2$, on the step $i = 3$ we have $4n + 4n = 8n$, i.e. $n2^2 + n2^2 = n2^3$, etc. It means that on the step $i = k$ we will have $n2^{k-1} + n2^{k-1} = n2^k$, that is why we can conclude that cardinality of resultant multiset of objects will be equal $n2^i$, where $i$ is recursion depth (step).□

**Theorem 5.** *Multiplicity of each object $A_j$ from multiset of objects $S$, which is obtained using RCL constructor, can be calculated by the following formula*

$$m(A_j) = 2^i,$$

*where $i$ is a recursion depth of RCL constructor.*

*Proof.* We know, that on each step $i$ RCL constructor will equally increase the multiplicity of all objects from set of objects $S_i$, it follows from the scheme of RCL constructor. According to Theorem 4, cardinality of resultant multiset of objects $|S| = n2^i$, where $i$ is recursion depth of RCL constructor. Combining these two facts, we can conclude that $m(A_j) = n2^i / n = 2^i$.□

## PS Constructor

First version of this constructor was presented in [Terletskyi, 2014] and now we are going to introduce its extension, which give us new abilities of its application. This constructor of multisets of objects is based on the idea of powerset of some set, which is why we will call it PS constructor.

**Example 14.** Let us consider again Example 12 and build all possible subsets of colors according to Definition 19 for set of colors $S = \{G, Y, R\}$, i.e.

$$S_1 = \{G,Y\},\ S_2 = \{G,R\},\ S_3 = \{Y,R\},\ S_4 = \{G,Y,R\}.$$

Let us apply union operation to sets of objects $S_1,\ldots,S_4$, i.e.

$$S = \{G,Y\}/T(S_1) \cup \{G,R\}/T(S_2) \cup \{Y,R\}/T(S_3) \cup \{G,Y,R\}/T(S_4) =$$
$$= \{G,Y,G,R,Y,R,G,Y,R\}/T(S),$$

where $S$ is a multiset of objects, and $T(S)$ is its class. However, such form of PS constructor does not provide indexation of light bulbs of the same color that is why we will improve it in this direction. As we can see, PS constructor consists of two parts, first of them is selection of subsets from basic set of objects, and second one is union of these subsets. Clearly, that we need to select all possible subsets of objects from basic set of objects in such way, that all copies of each object have unique index. That is why, we will organize selection procedure of subsets marking a choice of every object from set of objects $S$, during selection of every its subset using increase indexation of chosen objects from set of objects $S$ i.e.

$$S_1 = \{G,R\},\ S = \{Ind_1(G), Ind_1(Y), R\} = \{G_1, Y_1, R\};$$
$$S_2 = \{G_1, R\},\ S = \{Ind_1(G_1), Y_1, Ind_1(R)\} = \{G_2, Y_1, R_1\};$$
$$S_3 = \{Y_1, R_1\},\ S = \{G_2, Ind_1(Y_1), Ind_1(R_1)\} = \{G_2, Y_2, R_2\};$$
$$S_4 = \{G_2, Y_2, R_2\},\ S = \{Ind_1(G_2), Ind_1(Y_2), Ind_1(R_2)\} = \{G_3, Y_3, R_3\}.$$

Let us apply union operation to $S_1,\ldots,S_4$ and create new multiset of objects $S$, i.e.

$$S = \{G,Y\}/T(S_1) \cup \{G_1,R\}/T(S_2) \cup \{Y_1,R_1\}/T(S_3) \cup \{G_2,Y_2,R_2\}/T(S_4) =$$
$$= \{G,Y,G_1,R,Y_1,R_1,G_2,Y_2,R_2\}/T(S),$$

where $S$ is a multiset of objects, and $T(S)$ is its class. As the result we have obtained a multiset of objects, which consists of three objects $G$, $Y$, $R$ and their copies, which can be accurately identified and we can consider $S$ as one of possible projects of future electric garland. ♠

Now we can formulate and prove the following proposition.

**Proposition 1.** *The quantity of all possible subsets of sets of objects $S$ can be calculated by the following formula*

$$q(S_w) = 2^n - n - 1,$$

where $n = |S|$.

*Proof.* As we know, powerset of any set $A$ is the set of all subsets of $A$, including the empty set $\varnothing$ and $A$ itself, and it is denoted like

$$P(A) = \{P \mid P \subseteq A\}.$$

I.e. for the set $A = \{a,b,c\}$

$$P(A) = \{\{\varnothing\},\{a\},\{b\},\{c\},\{a,b\},\{a,c\},\{b,c\},\{a,b,c\}\}.$$

We also know that cardinality of powerset $|P(A)|$ of a set $A$ can be calculated using the following formula

$$|P(A)| = 2^n,$$

where $n = |A|$. However, according to the Definition 19, $\{\varnothing\}$, $\{a\}$, $\{b\}$, $\{c\}$ are not sets and set cannot be an element of another set. That is why in case of sets of objects previous formula can be rewritten as follows

$$q(S_w) = 2^n - n - 1,$$

where $S_w \subseteq S$, $q(S_w)$ is a quantity of all possible $S_w$. □

Using Proposition 1 and scheme of creation of multiset of objects from Example 14, we can represent our PS constructor as follows

$$PS(S) = \bigcup_{w=1}^{2^n - n - 1} S_w,$$

where $S$ is a basic set of objects for multiset of objects $PS(S)$ and $S_w \subseteq S$.

As we can see, PS constructor gives us determined scheme for creation of multiset of objects. We also can calculate multiplicity of every object and cardinality of multiset before its creation. As a proof of these facts, we can formulate and prove two following theorems.

**Theorem 6.** *The cardinality of each multiset of objects $S$, which is obtained using PS Constructor, can be calculated by the following formula*

$$|S| = \frac{n2^n}{2} - n,$$

where $n = |S_b|$.

*Proof.* Let us consider the set $S_1 = \{A, B, C\}$ and build a powerset for it, i.e.

$$P(S_1) = \{\{\varnothing\}, \{A\}, \{B\}, \{C\}, \{A,B\}, \{A,C\}, \{B,C\}, \{A,B,C\}\}.$$

Let us create the multiset $M_1$ as a union of all elements of $P(S_1)$, i.e.

$M_1 = \{\varnothing\} \cup \{A\} \cup \{B\} \cup \{C\} \cup \{A,B\} \cup \{A,C\} \cup \{B,C\} \cup \{A,B,C\} = \{A,B,C,A,B,A,C,B,C,A,B,C\}$.

Clearly that $|M_1| = 12$. Let us consider the set $S_2 = \{A, B, C, D\}$ and build a powerset for it, i.e.

$$P(S_2) = \{\{\varnothing\}, \{A\}, \{B\}, \{C\}, \{D\}, \{A,B\}, \{A,C\}, \{A,D\}, \{B,C\}, \{B,D\}, \{C,D\},$$
$$\{A,B,C\}, \{A,B,D\}, \{A,C,D\}, \{B,C,D\}, \{A,B,C,D\}\}.$$

Let us create the multiset $M_2$ as a union of all elements of $P(S_2)$, i.e.

$M_2 = \{\varnothing\} \cup \{A\} \cup \{B\} \cup \{C\} \cup \{D\} \cup \{A,B\} \cup \{A,C\} \cup \{A,D\} \cup \{B,C\} \cup \{B,D\} \cup \{C,D\} \cup$
$\cup \{A,B,C\} \cup \{A,B,D\} \cup \{A,C,D\} \cup \{B,C,D\} \cup \{A,B,C,D\} =$
$= \{A,B,C,D,A,B,A,C,A,D,B,C,B,D,C,D,A,B,C,A,B,D,A,C,D,B,C,D,A,B,C,D\}$.

As you can see $|M_2| = 32$. We know that $|P(S)| = 2^n$, where $n = |S|$. Clearly that in the case of $S_1$, $|P(S_1)| = 2^3 = 8$ and we can put

$$|M_1| = \frac{3 \cdot 2^3}{2} = 12,$$

as it is really so. In the case of $S_2$, $|P(S_2)| = 2^4 = 16$ and similar to the previous case we can put

$$|M_2| = \frac{4 \cdot 2^4}{2} = 32,$$

as we can see it is also true. Based on principle of mathematical induction we can conclude that for $P(S_n)$,

$$|M_n| = \frac{n 2^n}{2}.$$

Let us consider the set of objects $S_k = \{A, B, C\}$ and build for it all possible subsets of objects considering Definition 19, i.e.

$$S_1 = \{A, B\}, \ S_2 = \{A, C\}, \ S_3 = \{B, C\}, \ S_4 = \{A, B, C\}.$$

Let us create the multiset of objects $M_k$ as a union of all subset of $S_k$, i.e.

$$M_k = \{A, B\} / T(S_1) \cup \{A, C\} / T(S_2) \cup \{B, C\} / T(S_3) \cup \{A, B, C\} / T(S_4) =$$
$$= \{A, B, A, C, B, C, A, B, C\} / T(M_k).$$

Clearly that in the case of $|M_k|$, the formula which was used for calculation $|M_n|$ will be changed to

$$|M_k| = \frac{k 2^k}{2} - k,$$

where $k = |M_k|$. □

**Theorem 7.** *The multiplicity of each object $A_i$ from the multiset of objects S, which is obtained using PS constructor, can be calculated by the following formula*

$$m(A_i) = 2^{n-1} - 1,$$

where $n = |S_b|$.

*Proof.* We know that generating of possible subsets of objects $S_1, \ldots, S_w$ for set of objects $S_b$ can be represented as a combination of $k = \overline{2, n}$ different elements from the set of $n$ elements, i.e. $C_n^k$. During creation of subsets of cardinality $2$, we need to combine every object $A_i$ with every object from the set of objects $S_b \setminus A_i$. Clearly, we can create only $n-1$ such subsets, i.e. $C_{n-1}^1$. In the case of subsets of cardinality $3$, we will have $C_{n-1}^2$, and finally, in the case of subsets of cardinality $k$ we will have $C_{n-1}^{k-1}$.

According to the scheme of PC constructor, we can conclude that multiplicity of every object $A_i$ from multiset of objects $S$ consists of multiplicities of object $A_i$ in every subset of objects, i.e.

$$m(A_i) = \sum_{w=1}^{2^n-n-1} m_w(A_i),$$

where $m_w(A_i)$ is the multiplicity of object $A_i$ in the subset of objects $S_w \subseteq S_b$. It follows from $S_1 \cup ... \cup S_w$, where $w = 2^n - n - 1$. Using this fact, we can conclude that

$$m(A_i) = C_{n-1}^1 + ... + C_{n-1}^{n-1},$$

where $n = |S_b|$. It means that every object $A_i \in S$ has the same multiplicity. Using this fact, we can conclude that

$$m(A_i) = \frac{|S|}{|S_b|} = \frac{\frac{n2^n}{2} - n}{n} = \left(\frac{n2^n}{2} - n\right)\frac{1}{n} = \frac{n2^n}{2n} - \frac{n}{n} = \frac{2^n}{2} - 1 = 2^{n-1} - 1,$$

where $n = |S_b|$. □

Let us consider proof of Theorem 7. It shows that multiplicity of every object $A_i$ from multiset of objects $S$ can be calculated as a sum of appropriate binomial coefficients. Using this fact, we can build a part of Pascal's triangle. However, in contrast to original Pascal's triangle, we will combine its part with results of Theorem 6 and Theorem 7. It is convenient to formulate it as a following corollary.

**Corollary 7.1.** *We can calculate cardinality, multiplicity of every object from the multiset of objects, which was created using PS constructor, and quantity of subsets of objects which were used for its creation, using the following matrix*

| $m(A_k)$ | $\|S\|$ | $q(S_w)$ | $\|S_b\|$ | 2 | 3 | 4 | 5 | 6 | ... |
|---|---|---|---|---|---|---|---|---|---|
| 1 | 2 | 1 | 2 | 1 | | | | | |
| 3 | 9 | 4 | 3 | 3 | 1 | | | | |
| 7 | 28 | 11 | 4 | 6 | 4 | 1 | | | |
| 15 | 75 | 26 | 5 | 10 | 10 | 5 | 1 | | |
| 31 | 186 | 57 | 6 | 15 | 20 | 15 | 6 | 1 | |
| ... | ... | ... | ... | ... | ... | ... | ... | ... | ... |

,

where column $m(A_k)$ reflects multiplicity of object $A_k$ in multiset of objects $S$; column $|S|$ reflects cardinality of multiset of objects $S$; column $q(S_w)$ reflects quantity of $S_w \subseteq S_b$ that was used for establishing $S$; column $|S_b|$ reflects cardinality of basic sets of objects; first row starting with 5-th column reflects quantity of subsets of objects of certain cardinality, where cardinality coincides with the value of $a_{1, j \geq 5}$.

The elements of column $m(A_k)$ can be calculated using Theorem 7 or using the following formula

$$a_{i \geq 2, 1} = \begin{cases} 1, & i = 2; \\ \sum_{i \geq 2, j \geq 4} a_{i-1, j}, & i \geq 2. \end{cases}$$

The elements of column $|S|$ can be calculated using Theorem 6 or using the following formula

$$a_{i\geq 2,2} = \sum_{i\geq 2, j\geq 5} a_{i,j} \cdot a_{1,j}.$$

*The elements of column $q(S_w)$ can be calculated using Proposition 1 or using the following formula*

$$a_{i\geq 2,3} = \sum_{i\geq 2, j\geq 5} a_{i,j}.$$

*The element $a_{i\geq 2, j\geq 5}$ of the matrix can be calculated in such a way*

$$a_{i\geq 2, j\geq 5} = \begin{cases} 1, & j-i=1; \\ a_{i-1,j-1} + a_{i-1,j}, & j-i<3. \end{cases}$$

*or using the following formula*

$$a_{i\geq 2, j\geq 5} = \frac{a_{i,4}!}{a_{1,j}!(a_{i,4}-a_{1,j})!}.$$

## D2 Constructor

Similarly, to PS constructor, the first version of this constructor was also presented in [Terletskyi, 2014] and now we introduce its extension, which give us new abilities of its application. This constructor of multisets of objects is based on decomposition of basic set of objects on two disjoint subsets such, that in the result of their union we will obtain initial (basic) set of objects. That is why we call this constructor as D2 constructor.

**Example 15.** Let us consider Example 12 and imagine that we have light bulbs of green, yellow, red and blue colors, it means that we have set of colors $S=\{G,Y,R,B\}$. Let us perform D2 decomposition of it and find all possible variants of such decomposition, i.e.

$$S_1=\{G,Y\},\ S_2=\{R,B\};\ S_3=\{G,R\},\ S_4=\{Y,B\};\ S_5=\{G,B\},\ S_6=\{Y,R\}.$$

Let us apply union operation to these sets of objects, i.e.

$$S=\{G,Y\}/T(S_1)\cup\{R,B\}/T(S_2)\cup\{G,R\}/T(S_3)\cup\{Y,B\}/T(S_4)\cup$$
$$\cup\{G,B\}/T(S_5)\cup\{Y,R\}/T(S_6)=\{G,Y,R,B,G,R,Y,B,G,B,Y,R\}/T(S),$$

where $S$ is a multiset of objects, and $T(S)$ is its class. However, such form of D2 constructor does not provide indexation of light bulbs of same color that is why we will improve it in this direction.

As we can see, as the result of D2 decomposition of set of objects, we have obtained sets of objects $S_1,\ldots,S_6$. It means that in this case, there are three possible variants of such decomposition. Each variant of decomposition consists of pair of sets of objects. Let us change indexes of objects of these sets into accordance with number of decomposition's variant, using indexation operation, i.e.

$$S_1=\{Ind_1(G),Ind_1(Y)\}=\{G_1,Y_1\},\ S_2=\{Ind_1(R),Ind_1(B)\}=\{R_1,B_1\},$$
$$S_3=\{Ind_2(G),Ind_2(R)\}=\{G_2,R_2\},\ S_4=\{Ind_2(Y),Ind_2(B)\}=\{Y_2,B_2\},$$
$$S_5=\{Ind_3(G),Ind_3(B)\}=\{G_3,B_3\},\ S_6=\{Ind_3(Y),Ind_3(R)\}=\{Y_3,R_3\}.$$

Now, let us apply union operation to these sets and create new multiset of objects $S$, i.e.

$$S = \{G_1, Y_1\}/T(S_1) \cup \{R_1, B_1\}/T(S_2) \cup \{G_2, R_2\}/T(S_3) \cup \{Y_2, B_2\}/T(S_4) \cup \{G_3, B_3\}/T(S_5) \cup$$

$$\cup \{Y_3, R_3\}/T(S_6) = \{G_1, Y_1, R_1, B_1, G_2, R_2, Y_2, B_2, G_3, B_3, Y_3, R_3\}/T(S),$$

where $S$ is a multiset of objects, and $T(S)$ is its class. As the result we obtained multiset of objects $S$ which consists of four objects $G$, $Y$, $R$, $B$ and their copies, which can be accurately identified. We can consider $S$ as one of possible projects of future electric garland. ♠

Now we can formulate and prove the following proposition.

**Proposition 2.** *The quantity of all possible subsets of sets of objects $S$, which were obtained using D2 decomposition, can be calculated by the following formula*

$$q(S_w) = 2^n - 2n - 2,$$

where $n = |S|$.

*Proof.* From the previous section, we know that the quantity of all possible subsets of set of objects can be calculated as $q(S_w) = 2^n - n - 1$, where $n = |S|$. However, we can observe that the result of D2 decomposition of set of objects $S = \{G, Y, R, B\}$ does not contain subsets of cardinality $3$ and $4$, i.e. $n-1$ and $n$. It is true for any set of objects, because only sets of cardinality $n$ and $n-1$ cannot be divided according to principle of D2 decomposition. Clearly, that for each set of objects $S$ of cardinality $n$, only one subset of cardinality $n$ exists. Concerning subsets of cardinality $n-1$, their quantity can be calculated as

$$C_n^{n-1} = \frac{n!}{(n-1)!(n-(n-1))!} = \frac{n!}{(n-1)!1!} = \frac{n!}{(n-1)!} = n,$$

it follows from the proof of Theorem 7. Considering all these facts, we can conclude that

$$q(S_w) = 2^n - n - 1 - n - 1 = 2^n - 2n - 2,$$

where $n = |S|$. □

Using Proposition 2 and the scheme of creation of multiset of objects from Example 15, we can represent our D2 constructor as follows

$$D2(S) = \bigcup_{w=1}^{2^n - 2n - 2} (S_1 \cup S_2),$$

where $n = |S|$ and $S_1, S_2 \subseteq S$ are disjoint sets of objects, such that $S_1 \cup S_2 = S$.

As we can see, D2 constructor gives us determined scheme for creation of multiset of objects. We also can calculate multiplicity of every object and cardinality of the multiset before its creation. As a proof of these facts, we can formulate and prove two following theorems.

**Theorem 8.** *The cardinality of each multiset of objects $S$, which is obtained using D2 constructor, can be calculated by the following formula*

$$|S| = \frac{n2^n}{2} - n^2 - n,$$

where $n = |S_b|$.

**Proof.** According to Theorem 6, cardinality of each multiset of objects $S$, which is obtained using PS Constructor, can be calculated by the following formula

$$|S| = \frac{n2^n}{2} - n,$$

where $n = |S_b|$. From proof of Proposition 2, we know that result of D2 decomposition of set of objects $S$ does not contain subsets of cardinality $n-1$, $n$ and quantity of such subsets of objects will be equal $n$ and 1 respectively. That is why we can conclude that

$$|S| = \frac{n2^n}{2} - n(n-1) - n - n = \frac{n2^n}{2} - n^2 + n - 2n = \frac{n2^n}{2} - n^2 - n,$$

where $n = |S_b|$ □

**Theorem 9.** *Multiplicity of each object $A_i$ from multiset of objects $S$, which is obtained using D2 constructor, can be calculated by the following formula*

$$m(A_i) = 2^{n-1} - n - 1,$$

where $n = |S_b|$.

**Proof.** From proof of Theorem 7 we know that it is possible to build only $C_{n-1}^{k-1}$ subsets of cardinality $k$ for set of objects $S_b$, where $|S_b| = n$. In addition, we know that each object $A_i \in S$ has the same multiplicity. Using these facts, we can conclude that

$$m(A_i) = \frac{|S|}{|S_b|} = \frac{\frac{n2^n}{2} - n^2 - n}{n} = \left(\frac{n2^n}{2} - n^2 - n\right)\frac{1}{n} = \frac{n2^n}{2n} - \frac{n^2}{n} - \frac{n}{n} = \frac{2^n}{2} - n - 1 =$$

$$= \frac{2^n - 2n - 2}{2} = \frac{2(2^{n-1} - n - 1)}{2} = 2^{n-1} - n - 1,$$

where $n = |S_b|$. □

Let us consider proof of Theorem 9. It shows that multiplicity of every object $A_i$ from multiset of objects $S$ can be calculated as a sum of appropriate binomial coefficients. Using this fact, we can build a part of Pascal's triangle. However, in contrast to original Pascal's triangle, we will combine its part with results of Theorem 8 and Theorem 9. It is convenient to formulate this as following corollary.

**Corollary 9.1.** We can calculate cardinality, multiplicity of every object from the multiset of objects, which was created using D2 constructor, and quantity of subsets of objects which were used for its creation, using the following matrix

| $m(A_k)$ | $\|S\|$ | $q(S_w)$ | $\|S_b\|$ | 2 | 3 | 4 | 5 | 6 | ... |
|---|---|---|---|---|---|---|---|---|---|
| 3 | 12 | 6 | 4 | 6 | | | | | |
| 10 | 50 | 20 | 5 | 10 | 10 | | | | |
| 25 | 150 | 50 | 6 | 15 | 20 | 15 | | | |
| 56 | 392 | 112 | 7 | 21 | 35 | 35 | 21 | | |
| 119 | 952 | 238 | 8 | 28 | 56 | 70 | 56 | 28 | |
| ... | ... | ... | ... | ... | ... | ... | ... | ... | ... |

where column $m(A_k)$ reflects multiplicity of object $A_k$ in multiset of objects $S$; column $|S|$ reflects cardinality of multiset of objects $S$; column $q(S_w)$ reflects quantity of $S_w \subseteq S_b$ that was used for obtaining $S$; column $|S_b|$ reflects cardinality of basic sets of objects; first row starting with 5-th column reflects quantity of subsets of objects of certain cardinality, where cardinality coincides with the value of $a_{1,j \geq 5}$.

The elements of column $m(A_k)$ can be calculated using Theorem 9 or using the following formula

$$a_{i \geq 2,1} = \begin{cases} 3, & i = 2; \\ \sum_{i \geq 2, j \geq 4} a_{i-1,j}, & i > 2. \end{cases}$$

The elements of column $|S|$ can be calculated using Theorem 8 or using the following formula

$$a_{i \geq 2,2} = \sum_{i \geq 2, j \geq 5} a_{i,j} \cdot a_{1,j}.$$

The elements of column $q(S_w)$ can be calculated using Proposition 2 or using the following formula

$$a_{i \geq 2,3} = \sum_{i \geq 2, j \geq 5} a_{i,j}.$$

The element $a_{i \geq 2, j \geq 5}$ of matrix can be calculated in such a way

$$a_{i \geq 2, j \geq 5} = \begin{cases} 6, & i = 2, j = 5; \\ a_{i-1,j-1} + a_{1,j+1}, & j > 5, j - i = 3; \\ a_{i-1,j-1} + a_{i-1,j}, & j - i < 3; \end{cases}$$

or using the following formula

$$a_{i \geq 2, j \geq 5} = \frac{a_{i,4}!}{a_{1,j}!(a_{i,4} - a_{1,j})!}.$$

## Conclusions

This paper presents certain approach for modeling of some aspects of human thinking, in particular creation of sets and multisets of objects, within constructive object-oriented version of set theory, which was proposed in [Terletskyi, 2014]. The creation of sets and multisets of objects is considered from different sides, in particular classical set theory, object-oriented programming and development of intelligent information systems.

The paper also presents universal constructor of multisets of objects that gives us a possibility to create arbitrary multisets of objects and to recognize (identify) every copy of particular object, which have multiplicity $m \geq 2$. In addition, a few determined constructors of multisets of objects, which allow to create multisets, using strictly defined schemas, are also presented in the paper. The author proposed methods for calculation multiplicity of each object and cardinality of multiset before its creation for each constructor. That makes them very useful in cases of very big cardinalities of multisets.

The proposed approach for modeling of creation of sets and multisets of objects allows not only creation (generation) of sets and multisets of objects, but also their classification. It gives us an opportunity to consider the problem of object classification and identification in another way. The presented constructors of multisets of objects allow us to model corresponding processes of human thought, that in turn give us an opportunity to develop intelligent information systems, using these tools.

**Authors' Information**

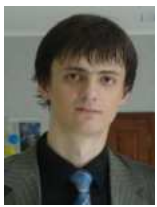

***Dmytro Terletskyi*** *– Postgraduate student, Department of Information Systems, Faculty of Cybernetics, Taras Shevchenko National University of Kyiv, 03680, 4d Glushkov Avenue, Kyiv, Ukraine; e-mail: dmytro.terletskyi@gmail.com*

*Major Fields of Scientific Research: Artificial Intelligence, Discrete Mathematics, Programming, Software Engineering.*